\documentclass[letterpaper]{article}

\usepackage[preprint]{aaai2027}
\usepackage{times}
\usepackage{helvet}
\usepackage{courier}
\usepackage[hyphens]{url}
\usepackage{graphicx}
\usepackage{natbib}
\usepackage{caption}
\usepackage{amsmath,amssymb,amsfonts}
\usepackage{bm}
\usepackage{booktabs}
\usepackage{multirow}
\usepackage{xcolor}
\usepackage{colortbl}
\usepackage{algorithm}
\usepackage{algorithmic}

\title{DAVET: Denoising-Aware Visual Evidence Trajectory Allocation \\ for Diffusion Vision-Language Models}

\author{
Yongkang Zhou\equalcontrib,
Xiang Xia\equalcontrib,
Cheng Yan,
Fan Xu,
Wuyang Zhang\corresponding
}
\affiliations{
University of Science and Technology of China\\
Hefei, China\\
yongkangz@mail.ustc.edu.cn, xxia@mail.ustc.edu.cn, wuyangz@ustc.edu.cn
}

\begin{document}
\maketitle

\begin{abstract}
Diffusion vision-language models (dVLMs) iteratively denoise masked responses while conditioning each denoising step on visual evidence, making visual conditioning a substantial recurring inference cost. Unlike autoregressive decoding, diffusion generation repeatedly revisits the entire response as uncertainty evolves. Our analysis reveals that visual evidence demand is strongly step-dependent, motivating adaptive allocation across denoising steps. Existing inference acceleration methods operate through decoding-side strategies or visual token compression via pruning and merging, but do not explicitly treat visual evidence as a resource whose demand evolves across the diffusion process.
Therefore, we present Denoising-Aware Visual Evidence Trajectory Allocation (DAVET), a training-free framework that allocates visual evidence according to the evolving generation state. Starting from a phase-conditioned evidence trajectory, the proposed allocation policy uses operation demand to set an evidence reserve whose allocation at each denoising step is modulated by trajectory risk. DAVET realizes the resulting budgets through a hierarchy of evidence views constructed from a single visual encoding, separating when and how much evidence is needed from how the evidence views are constructed.
Evaluated on two representative dVLMs, LLaDA-V and LaViDa, across multiple visual-understanding benchmarks, DAVET achieves an average speedup of 1.55$\times$ with an average relative performance drop of 1.86\%, showing that denoising-aware visual evidence allocation can reduce visual conditioning cost while largely preserving generation quality.
\end{abstract}

\section{Introduction}
\label{sec:intro}

Recently, diffusion language models (DLMs) are emerging as a promising alternative to autoregressive language models~\citep{d3pm,mdlm,blockdiffusion}. Rather than factorizing a sequence strictly from left to right, DLMs generate by iteratively denoising a corrupted sequence through bidirectional attention, allowing multiple positions to be refined in parallel and in flexible orders. LLaDA~\citep{llada} and Dream~\citep{dream} provide initial validation of this paradigm in large language models. LLaDA2.0~\citep{llada2} further extends this paradigm to 100B parameters, showing that DLMs can scale to substantially larger model capacities. Beyond general language modeling, DLMs are widely applied to complex reasoning~\citep{dot}, code generation~\citep{diffucoder}, and protein sequence design~\citep{dplm}.

Building on this progress, diffusion vision-language models (dVLMs), including LLaDA-V~\citep{lladav} and LaViDa~\citep{lavida}, extend DLMs to multimodal understanding by conditioning denoising on visual representations. Given an image and a query, a visual encoder produces visual tokens, while the diffusion language model conditions on these tokens to refine multiple answer positions in parallel through bidirectional attention. This extension introduces an efficiency bottleneck: high-resolution inputs often produce hundreds or thousands of visual tokens, far outnumbering the tokens in short answers, yet these visual representations participate in denoising computation at every step~\citep{lladafastv,d3tom}. Consequently, producing even a short answer incurs repeated computation over a long visual token sequence throughout the diffusion process, making visual conditioning a substantial cumulative burden, particularly for document, chart, and scene-text understanding.

\begin{figure}[t]
  \centering
  \includegraphics[width=\columnwidth]{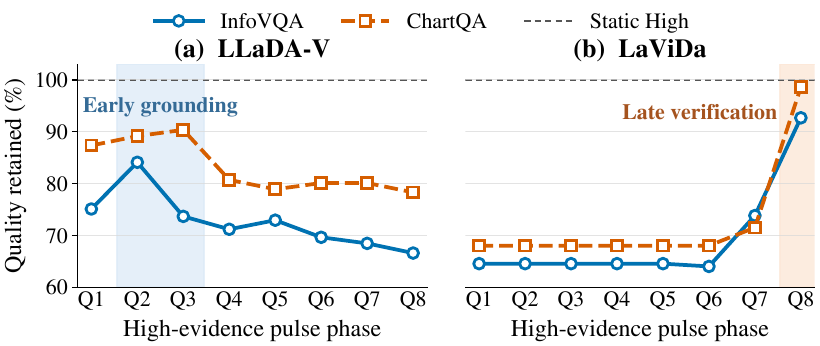}
  \caption{Backbone-dependent visual evidence demand across the diffusion process. With mixed evidence elsewhere, we shift a fixed-duration high-evidence pulse across eight equal denoising phases, from the earliest (Q1) to the latest (Q8). We report quality relative to full-evidence inference (Static High) on InfoVQA and ChartQA. LLaDA-V peaks early (Q2--Q3), whereas LaViDa peaks late (Q8), consistent with early grounding and late verification, respectively. Shading marks the corresponding peak regions.}
  \label{fig:phase_intervention}
\end{figure}

Crucially, visual evidence demand varies substantially at different denoising steps across the diffusion process. As shown in Figure~\ref{fig:phase_intervention}, we design controlled interventions that apply high-evidence conditioning for the same number of denoising steps during different phases of the diffusion process and observe markedly different generation quality. This evidence sensitivity is backbone-dependent: LLaDA-V is more sensitive to high-evidence conditioning during early denoising steps, whereas LaViDa is more sensitive during late denoising steps, patterns consistent with early grounding and late verification, respectively. Thus, the temporal pattern of visual evidence demand can differ across dVLM backbones. Efficient dVLM inference therefore depends not only on how much visual evidence is supplied, but also on when it is supplied across the diffusion process.

Yet existing dVLM inference acceleration methods are not designed around this step-dependent visual evidence demand. They primarily follow two directions. Decoding-side strategies accelerate generation through visual-token KV caching across denoising steps~\citep{fastdvlm} or by adapting which answer positions are committed in parallel~\citep{vrcd}. Visual token compression methods reduce visual conditioning cost by pruning~\citep{lladafastv,redvtp} or merging visual tokens~\citep{d3tom}. Decoding-side strategies do not determine when or how much visual evidence is supplied. Visual token compression methods can vary visual evidence across denoising steps, but define this variation through a particular pruning or merging rule. As a result, existing methods provide no general policy for allocating visual evidence according to the evolving generation state.

\noindent\textbf{Problem.} How can we allocate visual evidence across the diffusion process according to the evolving generation state, thereby reducing total visual conditioning cost while preserving generation quality, without retraining the model?

\noindent\textbf{Contribution.} Using controlled interventions that hold total high-evidence exposure constant, we reveal that visual evidence demand varies across denoising steps and exhibits backbone-specific early-grounding and late-verification patterns. Motivated by this finding, we formulate efficient visual conditioning in dVLMs as \emph{denoising-aware visual evidence allocation}, explicitly modeling when and how much visual evidence should be supplied as the generation state evolves. We then introduce Denoising-Aware Visual Evidence Trajectory Allocation (DAVET), a training-free framework that separates evidence allocation from evidence realization. Starting from a phase-conditioned evidence trajectory, the proposed allocation policy uses operation demand to set an evidence reserve whose allocation at each denoising step is modulated by trajectory risk. By separating evidence budgets from view construction, DAVET allows pooling, pruning, or merging to serve as evidence realizers without redefining the allocation policy. Empirically, we evaluate DAVET on two representative dVLMs, LLaDA-V and LaViDa, across multiple visual-understanding benchmarks. DAVET achieves a 1.55$\times$ average speedup with a 1.86\% average relative performance drop, yielding a favorable quality--efficiency trade-off compared with static-resolution, visual token pruning, and token merging methods. Further analyses show that the three allocation signals are complementary and that the policy remains effective across evidence realizers and nearby parameter settings. Together, these results support denoising-aware visual evidence allocation as a general framework for efficient dVLM inference.

\section{Related Work}
\label{sec:related}

\subsection{Diffusion Vision-Language Models}

Diffusion vision-language models (dVLMs) extend masked diffusion generation to multimodal inputs. LLaDA-V~\citep{lladav} equips LLaDA with a visual encoder and connector through visual instruction tuning, while LaViDa~\citep{lavida} introduces complementary masking and a prefix-oriented diffusion formulation for multimodal understanding. Beyond these, Dimple~\citep{dimple} combines autoregressive and diffusion training stages, whereas MMaDA~\citep{mmada} uses a unified diffusion architecture for text reasoning, multimodal understanding, and image generation. These models establish distinct architectures and denoising dynamics for multimodal diffusion. We focus on reducing visual conditioning cost in understanding-oriented dVLMs through training-free visual evidence allocation.

\subsection{Efficient Inference in dVLMs}

Existing dVLM inference acceleration methods primarily follow decoding-side strategies and visual token compression. Decoding-side strategies accelerate answer refinement through parallel commitment, masked-token truncation, or visual-token KV caching. Dimple~\citep{dimple} adapts how many answer positions are committed at each step; Sparse-LaViDa~\citep{sparselavida} truncates redundant masked tokens and uses register tokens as compact representations; and VRCD~\citep{vrcd} prioritizes answer positions with complementary visual grounding during parallel commitment. Fast-dVLM~\citep{fastdvlm} adopts block diffusion to enable visual-token KV caching across denoising steps and speculative block decoding. These methods accelerate masked-response updates, but do not determine when or how much visual evidence is supplied.

Visual token compression methods instead reduce visual conditioning cost by shortening the visual token sequence~\citep{fastv,pyramiddrop,sparsevlm}. LLaDA-FastV~\citep{lladafastv} prunes visual tokens during the first denoising step and reuses the reduced sequence thereafter. RedVTP~\citep{redvtp} estimates visual-token importance from masked-response signals for pruning, while D3ToM~\citep{d3tom} uses the evolving response to guide visual token merging across denoising steps. Although these methods differ in whether compression is fixed or dynamic, their decisions are defined by which tokens to retain, remove, or merge. DAVET instead allocates an evidence budget according to the evolving generation state. Pruning and merging can therefore serve as evidence realizers without defining the allocation policy itself.

\subsection{Adaptive Visual Computation in VLMs}

Adaptive visual computation varies visual inputs or visual token sequences by sample or generation state. CARES~\citep{cares} selects a sufficient image resolution for each image--query pair before encoding, while ResAdapt~\citep{resadapt} allocates input resolution across video frames. During autoregressive generation, DyRate~\citep{dyrate} progressively adjusts the visual-token compression rate, and DyVTE~\citep{dyvte} predicts when visual tokens can exit according to the generation state. VisionPulse~\citep{visionpulse} estimates a visual-token retention budget per step from visual attention mass and applies step-wise pruning. These methods reflect varying visual evidence demand, but use input selection or autoregressive decoding. They leave open how visual evidence should be allocated across denoising steps in the diffusion process, where multiple answer positions are refined in parallel and visual evidence demand can evolve differently across dVLM backbones.

\section{Preliminaries and Problem Formulation}
\label{sec:prelim}

\subsection{Diffusion Vision-Language Model Inference}
\label{sec:prelim-dmvlm}

Given an image $x$ and a query $q$, a visual encoder $E_{\phi}$ produces a visual token sequence $v=E_{\phi}(x)$. Let $y=(y_1,\ldots,y_L)$ denote an answer sequence of length $L$, and let $y^{(t)}$ be its state after denoising step $t$. A diffusion vision-language model~\citep{lladav,lavida} initializes every position in $y^{(0)}$ with the special mask token $\mathtt{[MASK]}$ and refines the sequence over $T$ steps. At step $t$, the unresolved positions are $\mathcal{M}_t=\{i\mid y_i^{(t-1)}=\mathtt{[MASK]}\}$. The model predicts their token distributions in parallel through bidirectional attention,
\begin{equation}
P_{t,i}=p_{\theta}\!\left(\,\cdot\mid y^{(t-1)},v,q\right),
\qquad i\in\mathcal{M}_t.
\label{eq:dvlm-update}
\end{equation}
Let $\hat{y}_{t,i}=\arg\max_{w}P_{t,i}(w)$, and let $\mathcal{U}_t\subseteq\mathcal{M}_t$ denote the positions selected for commitment by the decoding strategy.  The state is updated as
\begin{equation}
y_i^{(t)}=
\begin{cases}
\hat{y}_{t,i}, & i\in\mathcal{U}_t,\\
\mathtt{[MASK]}, & i\in\mathcal{M}_t\setminus\mathcal{U}_t,\\
y_i^{(t-1)}, & i\notin\mathcal{M}_t.
\end{cases}
\label{eq:dvlm-state-update}
\end{equation}
Repeating this update gradually resolves the masked positions until the final answer $\hat{y}=y^{(T)}$ is obtained.

Although $\mathcal{M}_t$ and the generation state evolve throughout the diffusion process, standard dVLM inference supplies the same $v$ each step. It repeatedly incurs the cost of conditioning on the unmodified visual sequence even when the current state may not require the same amount of visual evidence.

\subsection{Denoising-Aware Visual Evidence Allocation}
\label{sec:prelim-schedule}

Motivated by this mismatch, we treat visual evidence as a step-dependent resource rather than a fixed input. Let $s_t$ denote the generation state available before step $t$, comprising the denoising progress, query, current masked answer, and inference-time predictive statistics. Given a target average allocation budget $B$, an allocation policy $\pi$ determines the normalized budget at each step,
\begin{equation}
b_t=\pi(s_t;B)\in[0,1].
\label{eq:evidence-allocation}
\end{equation}
The sequence $\bm{b}_{1:T}=(b_1,\ldots,b_T)$ forms a visual evidence trajectory that specifies when and how much evidence is supplied across the diffusion process.

To separate visual evidence allocation from evidence realization, an evidence realizer $\mathcal{A}$ maps the encoded visual tokens $v$ and budget $b_t$ to the step-specific representation $v_t=\mathcal{A}(v;b_t)$. Here, $b_t=1$ leaves $v$ unchanged, while smaller budgets provide lower-granularity evidence at lower cost. This abstraction fixes neither the number of evidence levels nor their construction. Pooling, pruning, or merging can instantiate $\mathcal{A}$ through an ordered evidence--cost interface. Under $v_t$, Equation~(\ref{eq:dvlm-update}) becomes
$p_{\theta}(\,\cdot\mid y^{(t-1)},v_t,q)$.

Without changing the model or visual encoder parameters, the allocation problem is to preserve generation quality under a bounded average allocation budget:
\begin{equation}
\min_{\pi}\ 
\mathbb{E}\!\left[\mathcal{L}\!\left(\hat{y}_{\pi},y\right)\right]
\quad\mathrm{s.t.}\quad
\frac{1}{T}\sum_{t=1}^{T} b_t\le B,
\label{eq:evidence-objective}
\end{equation}
where $\hat{y}_{\pi}$ is the answer under policy $\pi$ and $\mathcal{L}$ is the task loss. The policy decides when and how much visual evidence is needed, whereas $\mathcal{A}$ determines how each budget level is represented. The normalized budget constrains allocation in the ordered evidence--cost space; exact latency depends on the realizer and hardware and is measured empirically.

This formulation also clarifies the scope of the problem. Static visual evidence corresponds to a constant trajectory $b_t=b$, while pooling, pruning, and merging can serve as different realizations of $\mathcal{A}$. Decoding-side strategies instead reduce the cost of updating the masked answer under a given $v_t$ and remain complementary to visual evidence allocation.

\section{The DAVET Framework}
\label{sec:method}

To account for step-dependent visual evidence demand during dVLM inference, we propose Denoising-Aware Visual Evidence Trajectory Allocation (DAVET), a training-free framework that allocates visual evidence according to the evolving generation state across the diffusion process. We first provide an overview, then detail the denoising-aware allocation policy and visual evidence realization.

\subsection{Overview}
\label{sec:method-overview}

DAVET comprises a visual evidence allocation policy and an evidence realizer. At denoising step $t$, its allocation policy specializes Equation~(\ref{eq:evidence-allocation}) using three complementary factors: the phase profile $\bm{p}_{1:T}$, operation demand $d(o)$ for operation category $o$, and trajectory risk $r_t$. The policy outputs a normalized evidence budget $b_t$, which identifies the corresponding step-specific representation $v_t$ among the evidence views constructed by $\mathcal{A}$:
\begin{equation}
b_t=\pi(\bm{p}_{1:T},d(o),r_t;B),\qquad
v_t=\mathcal{A}(v;b_t).
\label{eq:davet-overview}
\end{equation}
The policy $\pi$ determines \emph{when} and \emph{how much} visual evidence is supplied, whereas the evidence realizer $\mathcal{A}$ constructs evidence views for different budget levels before denoising. This separation allows the same allocation policy to operate with pooling, pruning, or merging as the underlying realizer.

\begin{figure*}[t]
  \centering
  \includegraphics[width=\textwidth]{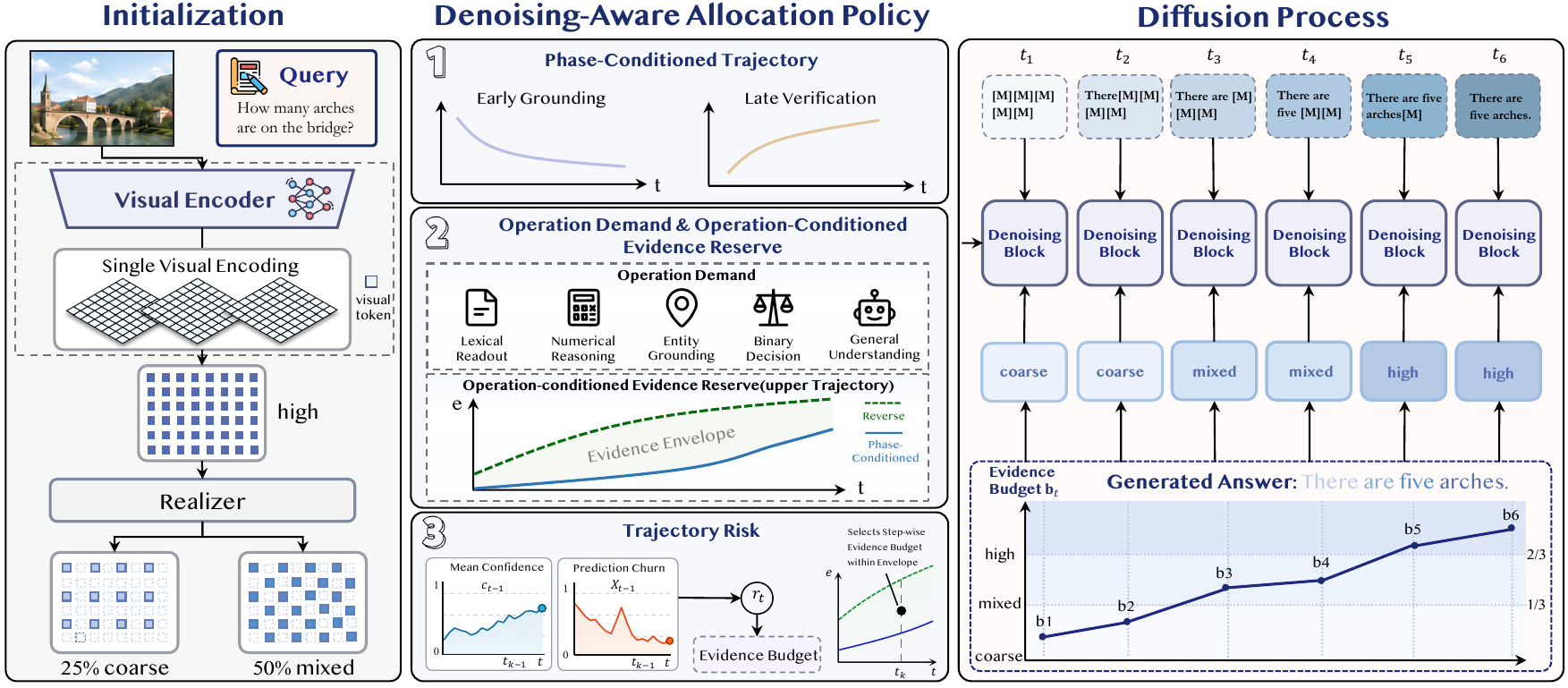}
  \caption{Overview of DAVET. During initialization, the image is encoded once, and the evidence realizer constructs an ordered hierarchy of evidence views from this representation. The allocation policy forms an evidence envelope whose lower trajectory follows the backbone-specific phase profile and whose upper trajectory adds a reserve determined by query-level operation demand. For each following denoising step, trajectory risk derived from mean confidence and prediction churn selects a budget within this envelope. The budget maps to an evidence view that conditions the parallel refinement of masked answer positions.}
  \label{fig:davet-overview}
\end{figure*}

Figure~\ref{fig:davet-overview} summarizes the allocation policy: the phase-conditioned trajectory distributes evidence across the diffusion process, operation demand bounds the reserve, and trajectory risk determines its step-wise use. The resulting budgets form a denoising-aware visual evidence trajectory.

\subsection{Denoising-Aware Allocation Policy}
\label{sec:method-policy}

DAVET represents the evidence budget at each denoising step as an interval, termed the \emph{evidence envelope}. The phase prior defines its lower trajectory, operation demand determines the reserve capacity and hence the upper trajectory, and trajectory risk selects the budget within these bounds. This factorization assigns each signal a distinct role: temporal structure, reserve capacity, or state-dependent utilization.

\noindent\textbf{Phase-conditioned evidence trajectory.}
As shown in Figure~\ref{fig:phase_intervention}, visual evidence demand varies across dVLM backbones over denoising time. We capture these patterns with a two-orientation phase family: $p_t=1-(t-0.5)/T$ for early grounding and $p_t=(t-0.5)/T$ for late verification. The $0.5$ offset centers each step in its interval, yielding symmetric profiles without zero endpoints. Accordingly, DAVET uses a backbone-specific orientation. The profile determines when visual evidence is emphasized, not its total budget. We use $\mathcal{N}(\bm{p}_{1:T};\cdot)$ to rescale the profile and clip it to $[0,1]$ so its mean matches the specified average budget. This normalization preserves phase ordering without manually assigning evidence levels to denoising windows.

\noindent\textbf{Operation-conditioned evidence reserve.}
The demand for fine-grained visual evidence varies across queries. Before denoising, DAVET captures this query-level prior with a shared training-free parser $g$. Using interrogative form and operation-indicative lexical cues, $g$ maps the query $q$ to a dominant operation category $o=g(q)$. The parser uses neither image content, dataset identity, nor model predictions. We use five categories---lexical readout, numerical reasoning, entity grounding, binary decision, and general understanding---each associated with a fixed demand prior $d(o)\in[0,1]$. The parser, taxonomy, and demand priors are shared across backbones and benchmarks. Appendix~A provides the cue rules, precedence, and prior specification.

Let $\rho\in[0,1]$ denote the maximum fraction of the average budget $B$ assigned to the operation-conditioned evidence reserve. The phase-conditioned lower trajectory and operation-conditioned upper trajectory are
\begin{equation}
\begin{aligned}
\underline{\bm{b}}
&=\mathcal{N}(\bm{p}_{1:T};(1-\rho)B),\\
\overline{\bm{b}}
&=\mathcal{N}(\bm{p}_{1:T};[1-\rho+\rho d(o)]B).
\end{aligned}
\label{eq:evidence-envelope}
\end{equation}
Their gap $\overline{\bm{b}}-\underline{\bm{b}}$ forms the operation-conditioned evidence reserve. Because both boundaries share $\bm{p}_{1:T}$, increasing $d(o)$ widens the evidence envelope without changing its phase ordering. The upper trajectory also has an average budget no greater than $B$. Thus, operation demand controls how much additional evidence is available, whereas trajectory risk determines how much is used at each step.

The deterministic parser is a lightweight training-free realization of the query-to-demand interface. Because DAVET consumes the bounded scalar $d(o)$, another demand estimator can replace $g$ without changing the allocation policy.

\noindent\textbf{Trajectory-risk modulation.}
The lower and upper trajectories define the evidence envelope at each denoising step, within which trajectory risk selects the final evidence budget as generation evolves. DAVET estimates this risk from two generation-state signals available during denoising. After step $t-1$, let $\bar{c}_{t-1}$ denote the mean confidence of provisional predictions over unresolved answer positions. We define $\chi_{t-1}$ as prediction churn: among answer positions that remain unresolved in both steps $t-2$ and $t-1$, it is the fraction whose highest-logit token changes between the two updates. For $t>1$, DAVET defines trajectory risk as
\begin{equation}
r_t = 1-\alpha \bar{c}_{t-1}(1-\chi_{t-1}),
\label{eq:trajectory-risk}
\end{equation}
where $\alpha\in[0,1]$ controls the influence of trajectory stability. The product $\bar{c}_{t-1}(1-\chi_{t-1})$ is high only when predictions are both confident and stable. Consequently, confident and stable predictions produce lower risk, whereas low confidence or high churn produces higher risk. Setting $\alpha=0$ gives $r_t=1$, while $\alpha=1$ applies the full stability signal.

Before any prediction state is available, DAVET sets $r_1=1$. When churn is unavailable after the first update, we set $\chi_1=0$. Given $r_t$, DAVET selects the final evidence budget by interpolating within the envelope:
\begin{equation}
b_t = \underline{b}_t
+ r_t\bigl(\overline{b}_t-\underline{b}_t\bigr).
\label{eq:envelope-allocation}
\end{equation}
Thus, $r_t=0$ selects the lower trajectory, while $r_t=1$ selects the upper trajectory. At each step $t>1$, DAVET uses the preceding update to compute risk and select the corresponding evidence view. These signals reuse model outputs, requiring neither an additional forward pass nor thresholded triggering. Since $r_t\in[0,1]$, selected budget remains inside the evidence envelope and its average does not exceed $B$.

\subsection{Visual Evidence Realization}
\label{sec:method-realizer}

\noindent\textbf{Ordered evidence hierarchy.}
Given the encoded visual sequence $v$, the evidence realizer $\mathcal{A}$ constructs $K$ ordered evidence views $\{v^{(1)},\ldots,v^{(K)}\}$ before denoising. Each view represents one evidence level, and the hierarchy increases in visual granularity and conditioning cost, with $v^{(K)}=v$. The number of levels controls how finely the continuous budget maps to visual evidence, but the allocation policy prescribes neither $K$ nor how the evidence views are constructed. Our implementation uses the special case $K=3$, yielding \texttt{coarse}, \texttt{mixed}, and \texttt{high} views from a single visual encoding. The \texttt{coarse} view applies structured pooling to preserve global coverage with fewer visual tokens. The \texttt{mixed} view augments this pooled representation with a fixed subset of unpooled tokens, restoring selected fine-grained details. The \texttt{high} view retains the uncompressed visual token sequence. All views are constructed before denoising and reused throughout the diffusion process.
A $K$-level hierarchy partitions the normalized budget range into $K$ ordered intervals, using canonical thirds for $K=3$:
\begin{equation}
v_t=
\begin{cases}
v^{\texttt{coarse}}, & b_t<\frac{1}{3},\\
v^{\texttt{mixed}}, & \frac{1}{3}\le b_t<\frac{2}{3},\\
v^{\texttt{high}}, & b_t\ge\frac{2}{3}.
\end{cases}
\label{eq:evidence-quantization}
\end{equation}
These thresholds define ordinal evidence levels. Equal-width budget intervals do not imply proportional FLOPs or latency. At each denoising step, $b_t$ selects among the constructed views without rerunning the visual encoder or reconstructing visual evidence. The hierarchy is ordered by visual sequence length, while generation latency is measured empirically.

\noindent\textbf{Compatible evidence realizers.}
Our three-level pooling construction is an evidence realizer, not a DAVET restriction. Any construction exposing an ordered evidence--cost hierarchy can instantiate $\mathcal{A}$: pruning varies retention ratios, whereas merging varies aggregation strength. More levels refine budget intervals without changing the allocation policy. Thus, pooling, pruning, and merging realize the same trajectory through different evidence views.

\section{Experiments}
\label{sec:exp}

\newcommand{\davgain}[1]{$_{{\color{green!45!black}\scriptstyle +#1\%}}$}
\newcommand{\davloss}[1]{$_{{\color{red!70!black}\scriptstyle -#1\%}}$}
\newcommand{\davspeed}[1]{$_{{\color{green!45!black}\scriptstyle \uparrow #1\times}}$}
\newcommand{\davslow}[1]{$_{{\color{red!70!black}\scriptstyle \downarrow #1\times}}$}
\newcommand{\davdefault}[1]{\cellcolor{blue!8}#1}

\begin{table*}[t]
\centering
{
\small
\setlength{\tabcolsep}{1mm}
\begin{tabular}{ccccccccc}
\toprule
\multirow{2}{*}{\textbf{Method}} &
\multicolumn{2}{c}{\textbf{InfoVQA}} &
\multicolumn{2}{c}{\textbf{ChartQA}} &
\multicolumn{2}{c}{\textbf{DocVQA}} &
\multicolumn{2}{c}{\textbf{TextVQA}} \\
\cmidrule(lr){2-3}\cmidrule(lr){4-5}\cmidrule(lr){6-7}\cmidrule(lr){8-9}
& Score $\Uparrow$ & Latency (s) $\Downarrow$
& Score $\Uparrow$ & Latency (s) $\Downarrow$
& Score $\Uparrow$ & Latency (s) $\Downarrow$
& Score $\Uparrow$ & Latency (s) $\Downarrow$ \\
\midrule
\rowcolor{gray!25}
\multicolumn{9}{c}{\textbf{LLaDA-V}} \\
\midrule
Static High
& 66.30 & 7.882
& 79.12 & 7.039
& 83.87 & 7.542
& 63.04 & 7.420 \\
Static Mid
& 65.15\davloss{1.73} & 7.501\davspeed{1.05}
& 73.96\davloss{6.52} & 4.023\davspeed{1.75}
& 84.59\davgain{0.86} & 7.702\davslow{0.98}
& 62.42\davloss{0.98} & 5.852\davspeed{1.27} \\
Static Low
& 50.91\davloss{23.21} & 5.366\davspeed{1.47}
& 38.28\davloss{51.62} & 2.529\davspeed{2.78}
& 79.82\davloss{4.83} & 6.076\davspeed{1.24}
& 52.53\davloss{16.67} & 2.597\davspeed{2.86} \\
\midrule
RedVTP
& 64.73\davloss{2.37} & 4.497\davspeed{1.75}
& 74.44\davloss{5.92} & 4.225\davspeed{1.67}
& 83.49\davloss{0.46} & 4.611\davspeed{1.64}
& 63.09\davgain{0.08} & 4.354\davspeed{1.70} \\
D3ToM
& 59.81\davloss{9.78} & 4.732\davspeed{1.67}
& 73.12\davloss{7.58} & 4.432\davspeed{1.59}
& 72.62\davloss{13.41} & 4.819\davspeed{1.57}
& 60.68\davloss{3.73} & 4.700\davspeed{1.58} \\
\midrule
\rowcolor{blue!8}
DAVET
& 65.57\davloss{1.10} & 5.551\davspeed{1.42}
& 78.36\davloss{0.96} & 5.277\davspeed{1.33}
& 83.18\davloss{0.82} & 5.665\davspeed{1.33}
& 62.86\davloss{0.29} & 5.464\davspeed{1.36} \\
\midrule
\rowcolor{gray!25}
\multicolumn{9}{c}{\textbf{LaViDa}} \\
\midrule
Static High
& 36.14 & 2.801
& 67.40 & 3.064
& 63.52 & 3.185
& 58.77 & 3.176 \\
Static Mid
& 35.72\davloss{1.15} & 2.782\davspeed{1.01}
& 61.80\davloss{8.31} & 2.282\davspeed{1.34}
& 63.62\davgain{0.16} & 3.152\davspeed{1.01}
& 58.71\davloss{0.11} & 3.110\davspeed{1.02} \\
Static Low
& 32.21\davloss{10.87} & 2.420\davspeed{1.16}
& 32.80\davloss{51.34} & 1.677\davspeed{1.83}
& 62.65\davloss{1.36} & 3.090\davspeed{1.03}
& 48.94\davloss{16.74} & 1.728\davspeed{1.84} \\
\midrule
RedVTP
& 31.14\davloss{13.84} & 2.280\davspeed{1.23}
& 52.12\davloss{22.67} & 2.350\davspeed{1.30}
& 53.15\davloss{16.32} & 2.428\davspeed{1.31}
& 54.60\davloss{7.10} & 2.483\davspeed{1.28} \\
D3ToM
& 33.61\davloss{7.00} & 2.350\davspeed{1.19}
& 57.56\davloss{14.60} & 2.543\davspeed{1.21}
& 55.18\davloss{13.12} & 2.665\davspeed{1.20}
& 57.38\davloss{2.37} & 2.588\davspeed{1.23} \\
\midrule
\rowcolor{blue!8}
DAVET
& 35.50\davloss{1.78} & 1.663\davspeed{1.68}
& 66.16\davloss{1.84} & 1.735\davspeed{1.77}
& 59.18\davloss{6.83} & 1.855\davspeed{1.72}
& 58.05\davloss{1.23} & 1.796\davspeed{1.77} \\
\bottomrule
\end{tabular}
}
\caption{Main results across two dVLM backbones and four benchmarks. Latency denotes average generation latency over all samples in each benchmark. Quality subscripts denote relative changes from Static High, while latency subscripts denote speedups. Green and red indicate favorable and unfavorable changes, respectively.}
\label{tab:main-results}
\end{table*}

We evaluate DAVET on two representative dVLMs across multiple visual-understanding benchmarks to determine whether aligning visual evidence with step-dependent demand improves the quality--efficiency trade-off of dVLM inference. This section presents the experimental setup, comparative results, a study of evidence realizers, component ablations, and hyper-parameter analysis. The experiments are organized around four main questions:

\begin{enumerate}
  \item[\textbf{Q1:}] Does DAVET improve the quality--efficiency trade-off across dVLM backbones and visual-understanding tasks?
  \item[\textbf{Q2:}] Can the same allocation policy remain effective with evidence realizers based on pooling, pruning, and merging?
  \item[\textbf{Q3:}] How does each allocation component contribute to the performance of DAVET across dVLM backbones?
  \item[\textbf{Q4:}] How does the behavior of DAVET change as its key hyper-parameters vary across dVLM backbones?
\end{enumerate}

\subsection{Experimental Settings}
\label{sec:exp-settings}

\noindent\textbf{Backbones, benchmarks, and metrics.}
We evaluate DAVET on two representative dVLMs, LLaDA-V~\citep{lladav} and LaViDa~\citep{lavida}. For both backbones, we use a generation length and block length of $32$. LLaDA-V and LaViDa use $16$ and $32$ denoising steps, respectively, and select answer positions for commitment according to confidence ranking. All experiments use deterministic decoding at temperature $0$ and run on NVIDIA A100 80GB GPUs. We evaluate on four widely used benchmarks including InfoVQA~\citep{infovqa}, ChartQA~\citep{chartqa}, DocVQA~\citep{docvqa}, and TextVQA~\citep{textvqa}. We report ANLS for InfoVQA and DocVQA, relaxed accuracy for ChartQA, and VQA accuracy with standard answer normalization for TextVQA. We report the average generation latency over all samples in each benchmark, excluding data loading and evaluation.

\noindent\textbf{Comparison methods.}
We include three static-resolution settings as direct efficiency controls. Static High encodes images at their original resolution. Static Mid and Static Low first resize each spatial dimension to approximately $2/3$ and $1/3$ of the original, respectively, and then use the same visual encoder. We compare DAVET with the visual-token pruning method RedVTP~\citep{redvtp} and the dynamic token-merging method D3ToM~\citep{d3tom}. We use official implementations of both methods where available. Since the D3ToM implementation supports LaViDa but not LLaDA-V, we adapt it to LLaDA-V following the method specification. To ensure a fair comparison, we apply the same visual-token retention ratio to both methods within each backbone, setting it to $0.50$ for LLaDA-V and $0.75$ for LaViDa.

\noindent\textbf{DAVET configuration.}
Both backbones use the same allocation-policy formulation, query parser, operation taxonomy, and $K{=}3$ ordered evidence hierarchy. The average budget $B$ sets the overall evidence scale across the diffusion process, the reserve fraction $\rho$ specifies the maximum share of $B$ assigned to the operation-conditioned reserve, and the risk strength $\alpha$ controls how strongly trajectory stability modulates reserve use. As shown in Figure~\ref{fig:phase_intervention}, the two backbones exhibit markedly different evidence-demand dynamics. We therefore align the phase orientation with each backbone: early grounding for LLaDA-V and late verification for LaViDa. Under these orientations, LLaDA-V uses $B{=}0.75$, $\rho{=}0.20$, and $\alpha{=}1.0$, whereas LaViDa uses $B{=}0.45$, $\rho{=}0.20$, and $\alpha{=}0.5$. Configurations remain fixed across benchmarks.

\subsection{Main Results}
\label{sec:exp-main}

\begin{figure}[t]
\centering
\includegraphics[width=\columnwidth]{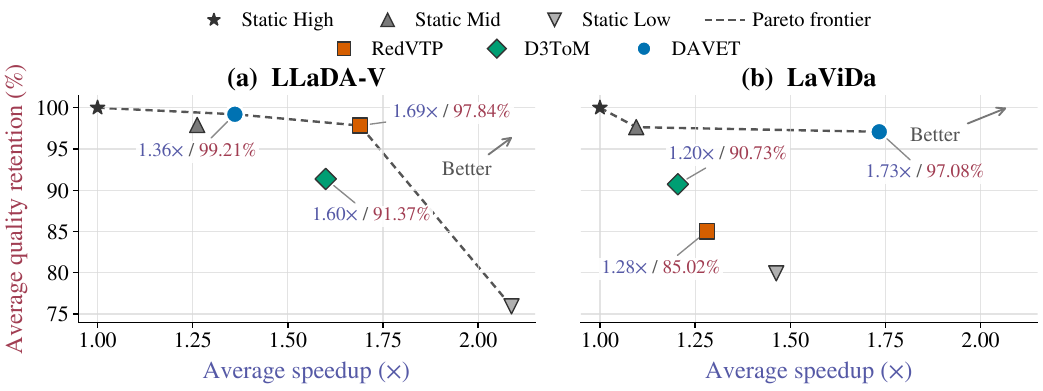}
\caption{Average quality--efficiency trade-offs across four benchmarks. For each method, the horizontal axis averages task-wise speedup over Static High, while the vertical axis averages quality retention. Dashed lines connect non-dominated points within each backbone. Annotations report speedup and quality retention; upper right is better.}
\label{fig:main-pareto}
\end{figure}

\noindent\textbf{To answer Q1.}
As shown in Table~\ref{tab:main-results}, the comparison methods exhibit strongly task-dependent quality--efficiency behavior. Static Low is fast but loses up to $51.62\%$ quality, whereas Static Mid preserves quality on some tasks but often provides little acceleration. RedVTP achieves strong quality--efficiency results on LLaDA-V, particularly on DocVQA and TextVQA, but incurs substantially larger quality losses on LaViDa. D3ToM also accelerates both backbones but exhibits broader quality degradation. These results show that visual-token compression is strongly backbone-dependent.

As shown in Figure~\ref{fig:main-pareto}, DAVET provides a more consistent quality-preserving trade-off. On LLaDA-V, it limits the relative quality drop to $0.29$--$1.10\%$ while achieving $1.33$--$1.42\times$ speedup, corresponding to $99.21\%$ average quality retention at $1.36\times$ average speedup. On LaViDa, DAVET achieves $97.08\%$ average quality retention at $1.73\times$ average speedup and outperforms the selected pruning and merging methods in both dimensions. Across all eight backbone--task pairs, DAVET achieves a $1.55\times$ average speedup with a $1.86\%$ average relative quality drop. The latency breakdown in Appendix~B further shows that evidence-view construction requires only $0.43$--$0.80$ ms per sample, while nearly all measured savings arise from reduced time spent on diffusion denoising. A reverse-orientation control in Appendix~C yields substantial quality loss at similar latency, supporting backbone-aligned evidence trajectories.

These results reflect the backbone-dependent evidence demand observed in Figure~\ref{fig:phase_intervention}. LLaDA-V's early-grounding trajectory preserves richer evidence near the beginning of denoising and yields a conservative but stable speedup. LaViDa's late-verification trajectory allows a larger portion of denoising to use reduced evidence before receiving finer-grained evidence near the end, producing greater acceleration. LaViDa DocVQA remains the main boundary, with a $6.83\%$ quality drop, suggesting that dense document understanding may require fine-grained evidence over a broader portion of the diffusion process. In a nutshell, although DAVET incurs its largest quality drop on LaViDa DocVQA, it still achieves a favorable average quality--efficiency trade-off across backbones and tasks, which answers Q1.

\subsection{Generality across Evidence Realizers}
\label{sec:exp-realizer}

\noindent\textbf{To answer Q2.}
We evaluate whether DAVET's allocation policy depends on a particular evidence realizer. We keep the policy fixed and construct the ordered evidence hierarchy through pooling, nested visual-token pruning, or hierarchical similarity merging, without introducing separate allocation rules. As shown in Table~\ref{tab:realizer}, all eight settings using pruning or merging remain within $0.51$--$1.60\%$ of Static High while achieving $1.25$--$1.78\times$ speedup. Quality remains tightly clustered among the three realizers within each backbone--task pair. Their latencies are also comparable, except that merging is moderately slower on LLaDA-V.

\begin{table}[t]
\centering
{
\small
\setlength{\tabcolsep}{0.6mm}
\begin{tabular}{ccccc}
\toprule
\multirow{2}{*}{\textbf{Realizer}} &
\multicolumn{2}{c}{\textbf{InfoVQA}} &
\multicolumn{2}{c}{\textbf{ChartQA}} \\
\cmidrule(lr){2-3}\cmidrule(lr){4-5}
& Score $\Uparrow$ & Latency (s) $\Downarrow$
& Score $\Uparrow$ & Latency (s) $\Downarrow$ \\
\midrule
\rowcolor{gray!25}
\multicolumn{5}{c}{\textbf{LLaDA-V}} \\
\midrule
\rowcolor{blue!8}
Pool
& 65.57\davloss{1.10} & 5.551\davspeed{1.42}
& 78.36\davloss{0.96} & 5.277\davspeed{1.33} \\
Prune
& 65.55\davloss{1.14} & 5.595\davspeed{1.41}
& 78.72\davloss{0.51} & 5.299\davspeed{1.33} \\
Merge
& 65.59\davloss{1.07} & 5.663\davspeed{1.39}
& 78.44\davloss{0.86} & 5.627\davspeed{1.25} \\
\midrule
\rowcolor{gray!25}
\multicolumn{5}{c}{\textbf{LaViDa}} \\
\midrule
\rowcolor{blue!8}
Pool
& 35.50\davloss{1.78} & 1.663\davspeed{1.68}
& 66.16\davloss{1.84} & 1.735\davspeed{1.77} \\
Prune
& 35.70\davloss{1.20} & 1.652\davspeed{1.70}
& 66.60\davloss{1.19} & 1.725\davspeed{1.78} \\
Merge
& 35.62\davloss{1.45} & 1.655\davspeed{1.69}
& 66.32\davloss{1.60} & 1.726\davspeed{1.78} \\
\bottomrule
\end{tabular}
}
\caption{Comparison of evidence realizers on InfoVQA and ChartQA. Pooling is the default. The allocation policy remains fixed, and only the construction of the ordered evidence hierarchy varies. Subscripts follow Table~\ref{tab:main-results}.}
\label{tab:realizer}
\end{table}

Despite different token contents, all hierarchies achieve similar quality under one allocation policy, while latency differences reflect construction costs. In a nutshell, the same allocation policy remains effective with evidence realizers based on pooling, pruning, and merging, which answers Q2.

\subsection{Ablation of Allocation Components}
\label{sec:exp-ablation}

\noindent\textbf{To answer Q3.}
We evaluate additive variants under the same evidence realizer. As shown in Table~\ref{tab:ablation}, Phase provides the largest initial efficiency gain but can sacrifice quality, particularly on LaViDa. Adding operation demand substantially recovers quality, while trajectory-risk modulation regains part of the efficiency with only minor score changes.

These changes reveal a functional decomposition rather than a uniform increase in visual evidence. Phase captures backbone-level temporal demand but applies the same trajectory across queries, explaining its strong efficiency and occasional quality loss. Operation demand exposes additional reserve capacity for queries likely to require fine-grained evidence, recovering quality at additional cost. Trajectory risk avoids using the full reserve when the evolving generation state is confident and stable, recovering part of this cost. Appendix~A further specifies a uniform-demand control with $d(o)=1$ that separates query-conditioned demand from uniform exposure to the maximum reserve. In a nutshell, phase provides temporal structure, operation demand provides query-conditioned reserve capacity, and trajectory risk controls state-dependent reserve use, which answers Q3.

\begin{table}[t]
\centering
{
\small
\setlength{\tabcolsep}{0mm}
\begin{tabular}{ccccc}
\toprule
\multirow{2}{*}{\textbf{Variant}} &
\multicolumn{2}{c}{\textbf{InfoVQA}} &
\multicolumn{2}{c}{\textbf{ChartQA}} \\
\cmidrule(lr){2-3}\cmidrule(lr){4-5}
& Score $\Uparrow$ & Latency (s) $\Downarrow$
& Score $\Uparrow$ & Latency (s) $\Downarrow$ \\
\midrule
\rowcolor{gray!25}
\multicolumn{5}{c}{\textbf{LLaDA-V}} \\
\midrule
Phase
& 64.64\davloss{2.51} & 4.940\davspeed{1.60}
& 78.08\davloss{1.31} & 4.638\davspeed{1.52} \\
$+$ Demand
& 65.59\davloss{1.06} & 5.754\davspeed{1.37}
& 78.32\davloss{1.01} & 5.386\davspeed{1.31} \\
\rowcolor{blue!8}
DAVET
& 65.57\davloss{1.10} & 5.551\davspeed{1.42}
& 78.36\davloss{0.96} & 5.277\davspeed{1.33} \\
\midrule
\rowcolor{gray!25}
\multicolumn{5}{c}{\textbf{LaViDa}} \\
\midrule
Phase
& 32.24\davloss{10.80} & 1.496\davspeed{1.87}
& 54.56\davloss{19.05} & 1.550\davspeed{1.98} \\
$+$ Demand
& 35.98\davloss{0.45} & 1.722\davspeed{1.63}
& 66.80\davloss{0.89} & 1.807\davspeed{1.70} \\
\rowcolor{blue!8}
DAVET
& 35.50\davloss{1.78} & 1.663\davspeed{1.68}
& 66.16\davloss{1.84} & 1.735\davspeed{1.77} \\
\bottomrule
\end{tabular}
}
\caption{Additive ablation of DAVET's allocation components on InfoVQA and ChartQA. Phase uses only the phase-conditioned evidence trajectory, $+$ Demand adds operation demand, and DAVET further adds trajectory-risk modulation. Subscripts follow Table~\ref{tab:main-results}.}
\label{tab:ablation}
\end{table}
\subsection{Hyper-parameter Analysis}
\label{sec:exp-sensitivity}

\noindent\textbf{To answer Q4.}
We conduct a hyper-parameter analysis of the average budget $B$, reserve fraction $\rho$, and risk strength $\alpha$ across both backbones. Detailed results and analysis are provided in Appendix~D. In a nutshell, these hyper-parameters induce interpretable quality--efficiency trade-offs around the backbone-specific defaults, with LaViDa showing greater sensitivity to aggressive settings, which answers Q4.

\section{Conclusion and Discussion}
\label{sec:conclusion}

\noindent\textbf{Conclusion.}
This work formulates efficient visual conditioning in dVLMs as denoising-aware visual evidence allocation. Motivated by backbone-specific early-grounding and late-verification patterns, we present DAVET, a training-free framework that derives an evidence envelope from a phase-conditioned trajectory and operation demand, then modulates the step-wise budget within this envelope according to trajectory risk. DAVET further separates this allocation process from evidence realization. Across LLaDA-V and LaViDa on four visual-understanding benchmarks, DAVET achieves a $1.55\times$ average speedup with a $1.86\%$ average relative performance drop. These results show that step-dependent visual evidence allocation can reduce recurring visual conditioning cost while largely preserving generation quality.

\noindent\textbf{Discussion.}
DAVET is an allocation framework rather than a visual-token compression method, allowing pooling, pruning, or merging to serve as evidence realizers. The deterministic parser is a lightweight query-to-demand interface and may be imperfect for ambiguous or compositional queries, but another demand estimator can replace it without changing the allocation policy. Our experiments instantiate the general $K$-level evidence hierarchy with $K{=}3$ and use one calibrated configuration per backbone across benchmarks. Automatic calibration, stronger demand estimation, and richer evidence hierarchies are natural directions for future work.

\bibliography{references}

\clearpage
\appendix
\setcounter{table}{0}
\renewcommand{\thetable}{\Alph{table}}
This appendix provides additional methodological details and supporting
analyses for DAVET. Appendix~A documents the operation parser and demand prior,
together with the uniform-demand control. Appendix~B decomposes inference
latency, Appendix~C evaluates phase orientation, and Appendix~D reports the
detailed hyper-parameter analysis.

\section{Operation Parser and Demand Prior}
\label{app:operation-parser}

Operation demand provides DAVET with a query-level prior on the amount of
fine-grained visual evidence that may be required. It is not a step-wise
allocation signal: the parser only determines the capacity of the
operation-conditioned reserve, while trajectory risk decides how much of that
reserve is used at each denoising step. This appendix documents the parser
interface, taxonomy, cue families, precedence, and demand prior used by DAVET.

\subsection{Parser Definition and Scope}

Given a query $q$, the deterministic parser $g$ returns one dominant operation
category $o=g(q)$. It uses only the query text, without accessing the image,
dataset identity, model predictions, or ground-truth answer. Before cue
matching, we lowercase and normalize the query and remove generic conversation
wrappers. The same parser, taxonomy, precedence, and demand prior are used for
both backbones on all benchmarks.

The taxonomy is intentionally compact. It distinguishes broad answer
operations that may differ in their demand for fine-grained visual evidence,
without introducing benchmark-specific labels. Importantly, the parser output
does not select an evidence view or a denoising step. It only controls the
width of the evidence reserve, leaving the phase profile and trajectory-risk
modulation unchanged.

\subsection{Operation Taxonomy}

Table~\ref{tab:operation-prior} defines five categories and a fixed bounded
demand prior, normalized by its largest value, to capture broad differences in
visual evidence demand.

\begin{table*}[t]
\centering
\small
\begin{tabular}{p{0.19\textwidth}p{0.57\textwidth}c}
\toprule
Operation category & Description & $d(o)$ \\
\midrule
Lexical readout &
Read fine text, labels, names, dates, or document fields. &
1.000 \\
Numerical reasoning &
Count, retrieve, compare, or compute numerical values. &
0.941 \\
Entity grounding &
Identify a person, place, organization, or visually grounded entity. &
0.824 \\
Binary decision &
Resolve a yes/no judgment grounded in the visual input. &
0.647 \\
General understanding &
Handle visual questions not matched by the preceding categories. &
0.529 \\
\bottomrule
\end{tabular}
\caption{Shared operation taxonomy and demand prior. The prior $d(o)$ controls
the operation-conditioned reserve capacity.}
\label{tab:operation-prior}
\end{table*}

The prior assigns larger reserves to lexical readout and numerical reasoning,
an intermediate reserve to entity grounding, and smaller reserves to binary
and general questions. These values are fixed across backbones and benchmarks.

\subsection{Cue Rules and Precedence}

The parser applies deterministic rules in order and returns the category of
the first match. Table~\ref{tab:operation-cues} summarizes the cue families.

\begin{table*}[t]
\centering
\small
\begin{tabular}{p{0.18\textwidth}p{0.72\textwidth}}
\toprule
Operation category & Cue groups \\
\midrule
Binary decision &
Yes/no interrogative forms, including auxiliary- and modal-initial queries. \\
Numerical reasoning &
Counting, percentage, arithmetic, comparison, and numerical-value requests. \\
Lexical readout &
Direct text-reading requests, dates, labels, names, titles, identifiers, and
document-field queries. \\
Entity grounding &
Person-, place-, organization-, and other entity-seeking interrogatives. \\
General understanding &
Choice and open-ended visual questions not captured above. \\
\bottomrule
\end{tabular}
\caption{Operation cues used by the shared parser. Phrases are matched after
query normalization.}
\label{tab:operation-cues}
\end{table*}

Rules proceed from explicit answer-form cues to broader operation cues and
return the first matching category. This fixed precedence resolves overlapping
cues consistently, while unmatched queries fall back to general understanding.

\subsection{Representative Examples}

Table~\ref{tab:operation-examples} illustrates representative mappings from
query forms to operation categories.

\begin{table}[t]
\centering
\small
\begin{tabular}{p{0.62\columnwidth}p{0.28\columnwidth}}
\toprule
Query & Operation category \\
\midrule
What is written on the sign? & Lexical readout \\
How many bars are shown? & Numerical reasoning \\
Who is shown in the photograph? & Entity grounding \\
Is the value increasing? & Binary decision \\
What is happening in the image? & General understanding \\
\bottomrule
\end{tabular}
\caption{Representative parser outputs.}
\label{tab:operation-examples}
\end{table}

\subsection{Role in DAVET}

The parser does not prescribe an evidence trajectory. Its output $d(o)$ only
sets the capacity of the operation-conditioned reserve:
\begin{equation}
\overline{\bm{b}}
=\mathcal{N}\!\left(
  \bm{p}_{1:T};
  [1-\rho+\rho d(o)]B
\right).
\end{equation}
Because the lower and upper trajectories share the same phase profile,
operation demand cannot change the backbone-specific phase ordering. Moreover,
the reserve is not automatically spent: trajectory risk selects the step-wise
budget within the evidence envelope according to the evolving generation
state. Operation demand can therefore widen or narrow the admissible reserve,
but it cannot directly relocate evidence across denoising steps.

This deterministic parser is a lightweight training-free realization of the
query-to-demand interface. DAVET consumes only the bounded scalar $d(o)$, so a
learned parser or another demand estimator can replace $g$ without changing the
allocation policy or evidence realizer.

\subsection{Uniform-Demand Control}

To separate query-conditioned demand from uniform exposure to the maximum
reserve, we construct a uniform-demand control that sets $d(o)=1$ for every
query. This control retains the phase orientation, average budget $B$, reserve
fraction $\rho$, risk strength $\alpha$, trajectory-risk modulation, evidence
realizer, budget thresholds, and generation protocol of DAVET. It therefore
exposes the maximum operation-conditioned reserve uniformly while leaving its
state-dependent use to trajectory risk. Comparison with the default demand
prior isolates the role of query-conditioned reserve capacity.

\begin{table}[t]
\centering
{
\small
\setlength{\tabcolsep}{0.45mm}
\begin{tabular}{llccc}
\toprule
\textbf{Setting} & \textbf{Demand} &
\textbf{Score} $\Uparrow$ & \textbf{Lat. (s)} $\Downarrow$ &
\textbf{High} \\
\midrule
\multirow{2}{*}{\shortstack[l]{LLaDA-V /\\InfoVQA}}
& Default & 65.57 & 5.551 & 9.41 \\
& Uniform & 65.66 & 5.726 & 9.85 \\
\cmidrule(lr){1-5}
\multirow{2}{*}{\shortstack[l]{LLaDA-V /\\ChartQA}}
& Default & 78.36 & 5.277 & 9.49 \\
& Uniform & 78.36 & 5.350 & 9.84 \\
\midrule
\multirow{2}{*}{\shortstack[l]{LaViDa /\\InfoVQA}}
& Default & 35.50 & 1.663 & 5.87 \\
& Uniform & 35.73 & 1.689 & 6.55 \\
\cmidrule(lr){1-5}
\multirow{2}{*}{\shortstack[l]{LaViDa /\\ChartQA}}
& Default & 66.16 & 1.735 & 5.83 \\
& Uniform & 66.92 & 1.761 & 6.35 \\
\bottomrule
\end{tabular}
}
\caption{Uniform-demand control on InfoVQA and ChartQA. Uniform sets $d(o)=1$
for every query, while Default uses the shared operation parser and demand
prior. High denotes the average number of denoising steps assigned high
evidence. All other DAVET components and settings remain fixed.}
\label{tab:uniform-demand}
\end{table}

As shown in Table~\ref{tab:uniform-demand}, exposing the maximum reserve to
every query increases average high-evidence use by $0.35$--$0.68$ denoising
steps and latency by $1.38$--$3.15\%$. The corresponding score changes range
from $0.00$ to $0.76$ points. Across the four settings, Uniform gains
$0.27$ score points on average while incurring a $1.90\%$ latency overhead.
Thus, query-conditioned demand reduces reserve exposure and latency while
differing from uniform maximum demand by $0.27$ score points on average.

\section{Latency Breakdown}
\label{app:latency-breakdown}

To identify the source of DAVET's acceleration, we profile Static High and
DAVET on matched subsets of InfoVQA and ChartQA. CUDA-synchronized timing
compares total latency and isolates evidence-view construction and diffusion
denoising.

The evidence-view component measures the one-time construction of the
\texttt{coarse} and \texttt{mixed} views from the encoded visual sequence.
The denoising component covers the repeated diffusion-model updates after
these views are available. End-to-end latency retains the remaining shared
inference overhead, allowing the decomposition to distinguish one-time view
construction from the recurring visual-conditioning cost targeted by DAVET.

\begin{table}[t]
\centering
{
\small
\setlength{\tabcolsep}{1.2mm}
\begin{tabular}{lccc}
\toprule
\multicolumn{4}{c}{\textbf{(a) End-to-end latency}} \\
\midrule
\textbf{Setting} & \textbf{High (s)} & \textbf{DAVET (s)} &
\textbf{Speedup} \\
\midrule
LLaDA-V / InfoVQA & 7.919 & 5.634 & 1.41$\times$ \\
LLaDA-V / ChartQA & 6.249 & 4.518 & 1.38$\times$ \\
LaViDa / InfoVQA  & 2.864 & 1.653 & 1.73$\times$ \\
LaViDa / ChartQA  & 2.837 & 1.610 & 1.76$\times$ \\
\midrule
\multicolumn{4}{c}{\textbf{(b) Denoising and view construction}} \\
\midrule
\textbf{Setting} & \textbf{High (s)} & \textbf{DAVET (s)} &
\textbf{Views (ms)} \\
\midrule
LLaDA-V / InfoVQA & 7.765 & 5.479 & 0.80 \\
LLaDA-V / ChartQA & 6.189 & 4.457 & 0.76 \\
LaViDa / InfoVQA  & 2.800 & 1.587 & 0.43 \\
LaViDa / ChartQA  & 2.773 & 1.546 & 0.43 \\
\bottomrule
\end{tabular}
}
\caption{CUDA-synchronized latency breakdown under matched generation
settings. Panel (a) compares total Static High and DAVET latency. Panel (b)
compares their diffusion denoising time and reports evidence-view construction
overhead (Views).}
\label{tab:latency-breakdown}
\end{table}

Table~\ref{tab:latency-breakdown} shows that total latency decreases by
$1.21$--$2.29$ s, while evidence-view construction adds only
$0.43$--$0.80$ ms per sample. The corresponding denoising reductions account
for nearly all end-to-end savings, with view construction remaining below
$0.03\%$ of total DAVET latency in every setting. The
$1.38$--$1.76\times$ profiled speedups therefore arise primarily from
conditioning the diffusion model on shorter visual token sequences across
denoising steps, rather than from changes outside denoising. CUDA
synchronization slightly increases absolute latency, so these measurements
support component attribution rather than replace the main-table latency.

\section{Phase-Orientation Control}
\label{app:phase-orientation}

To isolate the role of phase orientation, we reverse each backbone's canonical
orientation while keeping $B$, $\rho$, $\alpha$, operation demand, trajectory
risk, the evidence realizer, and the generation protocol fixed. This control
changes when visual evidence is emphasized without introducing a different
allocation rule.

\begin{table}[t]
\centering
{
\small
\setlength{\tabcolsep}{0.7mm}
\begin{tabular}{lcccc}
\toprule
\multirow{2}{*}{\textbf{Orientation}} &
\multicolumn{2}{c}{\textbf{InfoVQA}} &
\multicolumn{2}{c}{\textbf{ChartQA}} \\
\cmidrule(lr){2-3}\cmidrule(lr){4-5}
& Score $\Uparrow$ & Lat. (s) $\Downarrow$
& Score $\Uparrow$ & Lat. (s) $\Downarrow$ \\
\midrule
\rowcolor{gray!25}
\multicolumn{5}{c}{\textbf{LLaDA-V}} \\
\midrule
Canonical Early & 65.57 & 5.551 & 78.36 & 5.277 \\
Reversed Late & 49.56 & 5.653 & 57.92 & 5.309 \\
\midrule
\rowcolor{gray!25}
\multicolumn{5}{c}{\textbf{LaViDa}} \\
\midrule
Canonical Late & 35.50 & 1.663 & 66.16 & 1.735 \\
Reversed Early & 26.17 & 1.629 & 38.68 & 1.699 \\
\bottomrule
\end{tabular}
}
\caption{Phase-orientation control on both backbones. For each backbone,
Reversed swaps only the canonical phase orientation; all other DAVET settings
remain fixed.}
\label{tab:phase-orientation}
\end{table}

As shown in Table~\ref{tab:phase-orientation}, reversing the phase orientation
reduces quality by $16.01$ and $20.44$ points on LLaDA-V and by $9.33$ and
$27.48$ points on LaViDa for InfoVQA and ChartQA, respectively. Relative
quality decreases reach $24.41$--$26.08\%$ on LLaDA-V and
$26.29$--$41.54\%$ on LaViDa, while latency changes by at most $2.09\%$.
The similar latency indicates that this degradation is not explained by a
materially different average computational cost. Instead, it results from
placing the available evidence in the opposite portion of the diffusion
process. Together with the controlled phase interventions in the main paper,
this result supports aligning phase orientation with backbone-dependent
denoising dynamics.

\section{Detailed Hyper-parameter Analysis}
\label{app:hyperparameter-analysis}

This appendix expands the hyper-parameter analysis in the main paper. We vary
each scalar over five values on InfoVQA and ChartQA while holding the other
settings at their backbone-specific defaults. The three sweeps isolate the
overall evidence scale, the fraction exposed as conditional reserve, and the
strength of state-dependent reserve use. Tables~\ref{tab:sensitivity-budget}--\ref{tab:sensitivity-risk}
report the complete results. Default values are
highlighted in blue.

\noindent\textbf{Average budget $B$.}
The average budget controls the overall evidence scale across the diffusion
process. As shown in Table~\ref{tab:sensitivity-budget}, increasing $B$ raises
latency monotonically for both backbones, but its quality effect differs.
Across the complete LLaDA-V range, latency increases by about $50\%$, whereas
InfoVQA and ChartQA improve by only $2.22$ and $1.12$ points. This gradual
saturation shows diminishing quality gains as the evidence budget increases.

LaViDa exhibits a sharper transition. Increasing $B$ from $0.25$ to $0.45$
recovers $7.56$ points on InfoVQA and $22.96$ points on ChartQA, indicating
that the two smaller budgets undersupply evidence. Beyond the default,
increasing $B$ from $0.45$ to $0.65$ adds $21.23$--$23.17\%$ latency but only
$0.66$--$1.24$ quality points. Quality gains therefore diminish substantially
once $B$ reaches $0.45$.

\begin{table*}[t]
\centering
{
\small
\setlength{\tabcolsep}{2.7mm}
\begin{tabular}{lcccc}
\toprule
\multirow{2}{*}{\textbf{Backbone / $B$}} &
\multicolumn{2}{c}{\textbf{InfoVQA}} &
\multicolumn{2}{c}{\textbf{ChartQA}} \\
\cmidrule(lr){2-3}\cmidrule(lr){4-5}
& Score $\Uparrow$ & Latency (s) $\Downarrow$
& Score $\Uparrow$ & Latency (s) $\Downarrow$ \\
\midrule
\rowcolor{gray!25}
\multicolumn{5}{c}{\textbf{LLaDA-V}} \\
\midrule
$0.55$ & 63.87 & 4.488 & 77.72 & 4.195 \\
$0.65$ & 64.71 & 5.045 & 78.12 & 4.745 \\
\davdefault{$0.75$} & \davdefault{65.57} & \davdefault{5.551} & \davdefault{78.36} & \davdefault{5.277} \\
$0.85$ & 65.78 & 6.156 & 78.52 & 5.788 \\
$0.95$ & 66.09 & 6.722 & 78.84 & 6.298 \\
\midrule
\rowcolor{gray!25}
\multicolumn{5}{c}{\textbf{LaViDa}} \\
\midrule
$0.25$ & 27.94 & 1.324 & 43.20 & 1.360 \\
$0.35$ & 28.62 & 1.414 & 43.72 & 1.440 \\
\davdefault{$0.45$} & \davdefault{35.50} & \davdefault{1.663} & \davdefault{66.16} & \davdefault{1.735} \\
$0.55$ & 36.16 & 1.855 & 67.32 & 1.947 \\
$0.65$ & 36.16 & 2.016 & 67.40 & 2.137 \\
\bottomrule
\end{tabular}
}
\caption{Sensitivity to the average budget $B$. The other DAVET
hyper-parameters remain at their backbone-specific defaults. Blue cells mark
the default value and its corresponding results.}
\label{tab:sensitivity-budget}
\end{table*}

\noindent\textbf{Reserve fraction $\rho$.}
The reserve fraction determines how much of $B$ is moved from the guaranteed
lower trajectory into the operation-conditioned reserve. A larger $\rho$
therefore makes a greater share of evidence conditional on both operation
demand and trajectory risk, allowing stable states to use lower budgets.
Table~\ref{tab:sensitivity-reserve} confirms this behavior: latency decreases
monotonically as $\rho$ grows on both backbones.

LLaDA-V remains stable across the full sweep. Moving from $\rho{=}0$ to
$0.40$ reduces latency by $11.21$--$12.02\%$, with quality changes of only
$0.20$--$0.91$ points. LaViDa InfoVQA follows a similar, though less stable,
trend. Its ChartQA quality, however, falls from $66.16$ at the default to
$63.48$ and $58.88$ at $\rho{=}0.30$ and $0.40$. At $\rho{=}0.20$, both
backbones remain close to lower-$\rho$ quality, whereas larger values expose
stronger sensitivity on LaViDa ChartQA.

\begin{table*}[t]
\centering
{
\small
\setlength{\tabcolsep}{2.7mm}
\begin{tabular}{lcccc}
\toprule
\multirow{2}{*}{\textbf{Backbone / $\rho$}} &
\multicolumn{2}{c}{\textbf{InfoVQA}} &
\multicolumn{2}{c}{\textbf{ChartQA}} \\
\cmidrule(lr){2-3}\cmidrule(lr){4-5}
& Score $\Uparrow$ & Latency (s) $\Downarrow$
& Score $\Uparrow$ & Latency (s) $\Downarrow$ \\
\midrule
\rowcolor{gray!25}
\multicolumn{5}{c}{\textbf{LLaDA-V}} \\
\midrule
$0.00$ & 65.87 & 6.014 & 78.44 & 5.639 \\
$0.10$ & 65.70 & 5.768 & 78.40 & 5.399 \\
\davdefault{$0.20$} & \davdefault{65.57} & \davdefault{5.551} & \davdefault{78.36} & \davdefault{5.277} \\
$0.30$ & 65.40 & 5.436 & 78.32 & 5.127 \\
$0.40$ & 64.96 & 5.291 & 78.24 & 5.007 \\
\midrule
\rowcolor{gray!25}
\multicolumn{5}{c}{\textbf{LaViDa}} \\
\midrule
$0.00$ & 36.01 & 1.735 & 67.08 & 1.824 \\
$0.10$ & 35.92 & 1.714 & 66.96 & 1.796 \\
\davdefault{$0.20$} & \davdefault{35.50} & \davdefault{1.663} & \davdefault{66.16} & \davdefault{1.735} \\
$0.30$ & 35.09 & 1.602 & 63.48 & 1.662 \\
$0.40$ & 34.07 & 1.549 & 58.88 & 1.592 \\
\bottomrule
\end{tabular}
}
\caption{Sensitivity to the reserve fraction $\rho$. The average budget and
risk strength remain at their backbone-specific defaults. Blue cells mark the
default value and its corresponding results.}
\label{tab:sensitivity-reserve}
\end{table*}

\noindent\textbf{Risk strength $\alpha$.}
The risk strength scales how strongly confidence and prediction churn modulate
reserve use. Larger $\alpha$ removes more reserve when the trajectory is
confident and stable. Table~\ref{tab:sensitivity-risk} accordingly shows
monotonic latency reductions as $\alpha$ increases.

LLaDA-V is nearly invariant across the sweep: moving from $\alpha{=}0$ to
$1$ changes quality by at most $0.04$ points and reduces latency by only
$1.80$--$2.48\%$. This limited sensitivity suggests that risk modulation has
a small aggregate effect under the tested LLaDA-V configuration. LaViDa
responds more strongly. Over the same range, latency falls by
$7.54$--$9.24\%$, while InfoVQA decreases by $0.67$ points and ChartQA by
$2.92$ points. The largest ChartQA reductions appear at $\alpha\geq0.75$.

Taken together, the sweeps show that the three hyper-parameters govern
different aspects of DAVET behavior. $B$ controls the global evidence scale,
$\rho$ determines how much evidence becomes conditional, and $\alpha$ controls
how aggressively the evolving generation state withdraws that conditional
reserve. Their effects are directionally interpretable, while low $B$, high
$\rho$, and high $\alpha$ expose clear quality boundaries, especially for
LaViDa.

\begin{table*}[t]
\centering
{
\small
\setlength{\tabcolsep}{2.7mm}
\begin{tabular}{lcccc}
\toprule
\multirow{2}{*}{\textbf{Backbone / $\alpha$}} &
\multicolumn{2}{c}{\textbf{InfoVQA}} &
\multicolumn{2}{c}{\textbf{ChartQA}} \\
\cmidrule(lr){2-3}\cmidrule(lr){4-5}
& Score $\Uparrow$ & Latency (s) $\Downarrow$
& Score $\Uparrow$ & Latency (s) $\Downarrow$ \\
\midrule
\rowcolor{gray!25}
\multicolumn{5}{c}{\textbf{LLaDA-V}} \\
\midrule
$0.00$ & 65.59 & 5.692 & 78.32 & 5.374 \\
$0.25$ & 65.59 & 5.670 & 78.28 & 5.338 \\
$0.50$ & 65.59 & 5.648 & 78.36 & 5.307 \\
$0.75$ & 65.57 & 5.617 & 78.32 & 5.272 \\
\davdefault{$1.00$} & \davdefault{65.57} & \davdefault{5.551} & \davdefault{78.36} & \davdefault{5.277} \\
\midrule
\rowcolor{gray!25}
\multicolumn{5}{c}{\textbf{LaViDa}} \\
\midrule
$0.00$ & 35.98 & 1.710 & 66.80 & 1.818 \\
$0.25$ & 35.84 & 1.673 & 66.24 & 1.775 \\
\davdefault{$0.50$} & \davdefault{35.50} & \davdefault{1.663} & \davdefault{66.16} & \davdefault{1.735} \\
$0.75$ & 35.52 & 1.613 & 65.36 & 1.695 \\
$1.00$ & 35.31 & 1.581 & 63.88 & 1.650 \\
\bottomrule
\end{tabular}
}
\caption{Sensitivity to the risk strength $\alpha$. The average budget and
reserve fraction remain at their backbone-specific defaults. Blue cells mark
the default value and its corresponding results.}
\label{tab:sensitivity-risk}
\end{table*}

\end{document}